\documentclass[letterpaper, 10 pt, conference]{ieeeconf}  

\usepackage{svg}
\usepackage{graphicx}
\usepackage{amsmath}
\usepackage{mathtools}
\usepackage{booktabs}
\usepackage{multirow}
\usepackage{cuted}
\usepackage{caption}
\usepackage{amssymb}
\usepackage{hyperref}

\IEEEoverridecommandlockouts                              

\title{\LARGE \bf
IN-Sight: Interactive Navigation through Sight
}

\author{
    Philipp Schoch, Fan Yang, Yuntao Ma, Stefan Leutenegger, Marco Hutter and Quentin Leboutet%
    \thanks{*This work was supported by Intel Labs Munich.}
    \thanks{Philipp Schoch, Fan Yang, Yuntao Ma, and Marco Hutter are with the Robotic Systems Lab, ETH, 8092 Zürich, Switzerland. Email: {\tt\small pschoc@student.ethz.ch; fanyang1@ethz.ch; mayun@ethz.ch; mahutter@ethz.ch}.  }%
    \thanks{Stefan Leutenegger is with the Smart Robotics Lab, TUM, 85748 Garching, Germany. Email: {\tt\small Stefan.Leutenegger@tum.de}.}%
    \thanks{Quentin Leboutet is with Intel Germany, 85579 Neubiberg, Germany. Email: {\tt\small quentin.leboutet@intel.com}. }%
}

\begin{document}

\maketitle
\thispagestyle{empty}
\pagestyle{empty}


\begin{abstract}

Current visual navigation systems often treat the environment as static, lacking the ability to adaptively interact with obstacles. This limitation leads to navigation failure when encountering unavoidable obstructions. In response, we introduce IN-Sight, a novel approach to self-supervised path planning, enabling more effective navigation strategies through interaction with obstacles. Utilizing RGB-D observations, IN-Sight calculates traversability scores and incorporates them into a semantic map, facilitating long-range path planning in complex, maze-like environments. To precisely navigate around obstacles, IN-Sight employs a local planner, trained imperatively on a differentiable costmap using representation learning techniques. The entire framework undergoes end-to-end training within the state-of-the-art photorealistic Intel SPEAR Simulator. We validate the effectiveness of IN-Sight through extensive benchmarking in a variety of simulated scenarios and ablation studies. Moreover, we demonstrate the system's real-world applicability with zero-shot sim-to-real transfer, deploying our planner on the legged robot platform ANYmal, showcasing its practical potential for interactive navigation in real environments. \\
Supplementary Video: \url{https://youtu.be/ja0Vjm72ZDw}

\end{abstract}

\section{INTRODUCTION}

\begin{figure}
    \centering
    \includegraphics[width=1\linewidth]{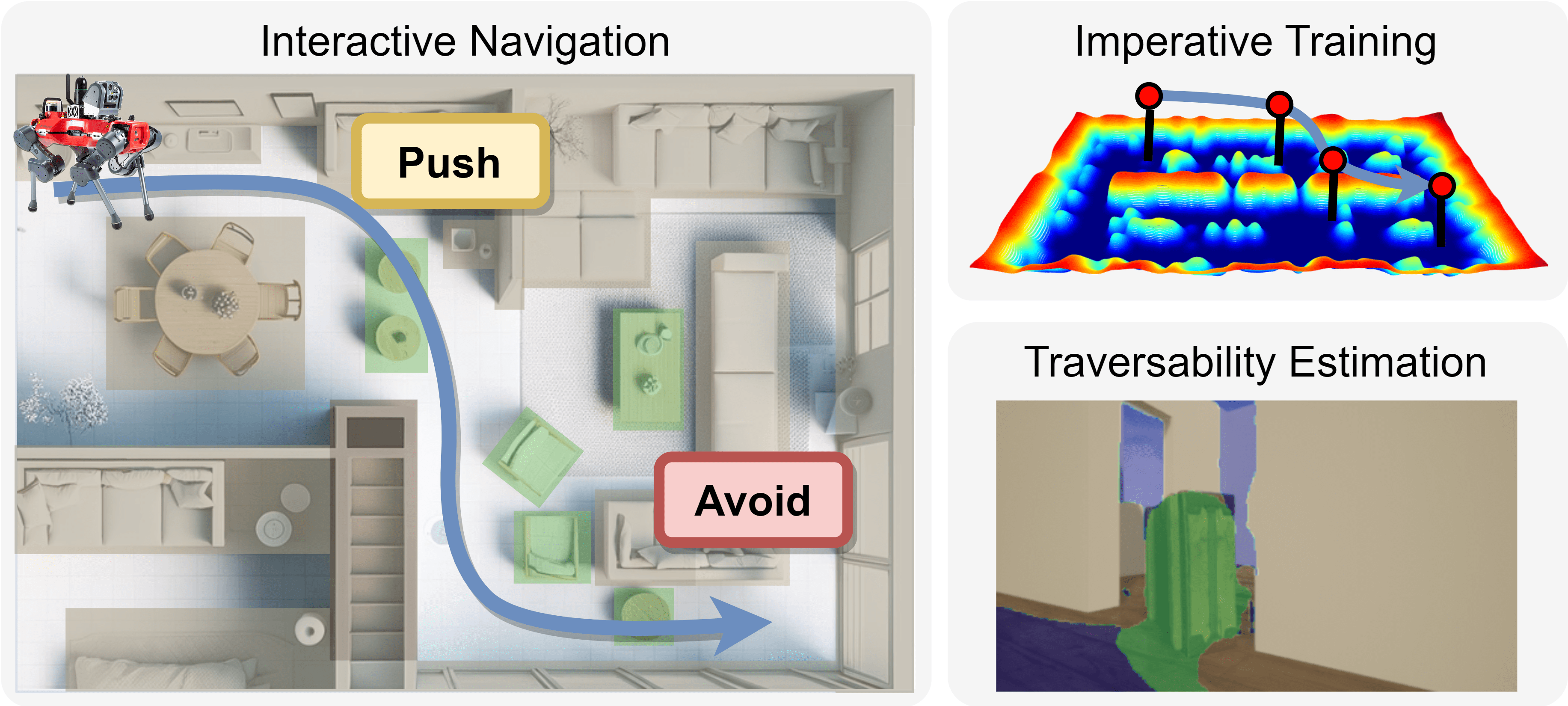}
    \caption{Interactive Navigation: while traversing to the goal, the agent avoids stationary obstructions (brown) and pushes light obstacles (green) out of the way. Imperative Training: A self-supervised training methodology is used to train the agent end-to-end on a differentiable costmap. Traversability Estimation (TE): the agent learns to distinguish between traversable terrain (blue/grey), interactive objects (green) and static obstacles (brown) by solving TE as a co-task. The traversability estimates are integrated into a map for long-horizon path planning.}
    \label{fig:method_overview}
\end{figure}

In robotics, navigating real-world environments from vision sensors has been a long studied subject \cite{yasuda2020autonomous, zhu2021deep, khan2022recent, arafat2023vision}. Yet, most traditional navigation approaches treat the environment as a passive entity, not capturing the dynamic nature of the real world where obstacles can be moved deliberately. Visual Interactive Navigation (VIN) takes this into account by engaging with the surrounding, actively moving obstacles out of the way to enable more effective traversal strategies (cf. Figure \ref{fig:method_overview}).

Engaging with the environment introduces several challenges. Firstly, an agent must discern whether an obstacle is movable and whether moving it is beneficial for reaching the goal. This involves predicting how obstacles will respond to interaction, a task complicated by the intricate nature of such dynamics. As a result, the field gravitates towards learning-based approaches. Reinforcement Learning (RL) has shown promise in enabling navigation in interactive indoor environments using vision sensors, but it can suffer from poor sample efficiency, especially with high-resolution visual inputs. Therefore, many studies resort to fast non-photorealistic physics simulators \cite{interactiveNav:Beyond-Tabula-Rasa}\cite{interactiveNav:pushingOutOfWay}\cite{interactiveNav:RelMoGen} or vector observations rather than raw visual inputs \cite{interactiveNav:HRLForInteractiveNav}. While successful in simulation, these approaches often fall short in real-world application.

Real-world environments, characterized by maze-like structures such as buildings or public spaces, pose a significant challenge to agents that rely on purely reactive policies without utilizing past observations, leading them to easily get lost. \cite{interactiveNav:Transformer_Memory} attempts to mitigate this by integrating a transformer-based memory into the RL policy, however causing stability complications during training and limited sim2real transfer. In contrast, the supervised learning approach \cite{Nav:NAMO} incorporates memory using explicit mapping. The authors fine-tune a pretrained object pose estimation module to localize paper boxes, combining it with explicit path planning on a pre-computed map. This approach alleviates the memory problem and avoids training on non-photorealistic data, improving applicability to real-world environments. However, this work does neither address the complexities related to reasoning whether an obstacle can be moved, nor how generalization to unseen obstacles with similar properties can be achieved. Relying on a trial-and-error approach, the robot has to actively push the obstacle to determine whether its movable, resulting in many undesired interactions. Given that the visual appearance of real-world environments varies drastically and physical sensors are affected by sensing artifacts, deploying VIN systems to robotic hardware remains a major complication.

In this work we introduce a hierarchical planner, continuously integrating traversability estimates into a global map, which is then used to perform high-level path planning. From the high-level path plan we extract subgoals, which are then tracked by a local planner. This allows us to navigate cluttered and maze-like environments at scale. Contrary to previous RL-based approaches, we train our system in a self-supervised fashion, achieving fast convergence to a robust policy. We generate a diverse photorealistic training set consisting of sequential data using the Intel SPEAR simulator. We further employ various robustification techniques in order to mitigate the effects of sensing artifacts and imperfect photorealism.


Our main contributions are as follows:

\begin{itemize}
    \item We propose a new training paradigm by extending the established imperative approach \cite{Nav:Iplanner} to accommodate interactive environments, leveraging sequential observations to support interactive navigation.
    \item We introduce a scalable approach to generate self-supervision for path prediction and traversability estimation, based on different heuristics such as obstacle mass, friction, and terrain inclination.
    \item We employ a blend of robustification strategies and accelerate the ICL-NIUM depth noise algorithm \cite{aux:depth_sensor_model} by two orders of magnitude, allowing to simulate truly random depth sensor artifacts without slowing down training.
    \item We demonstrate successful zero-shot transfer of the learned policy to the ANYmal legged robot.
\end{itemize}

\section{RELATED WORK}
Visual navigation is a widely explored research field with a substantial amount of influential contributions over the past two decades. A variety of relevant concepts, developed to address the different challenges of visual navigation, can be directly applied to visual interactive navigation. 
As mentioned previously, memory allows the agent to utilize past observations in decision processes. Not only limited to classical planning pipelines, memory also becomes an essential building block for successful learning-based architectures \cite{Nav:NAMO}. Often encountered in hierarchical planners, mapping allows to leverage explicit path planning methods such as A-star, Dijkstra's algorithm, or fast marching methods \cite{aux:FMMs}. Beyond simple occupancy maps used in classical pipelines, learning-based approaches enable to encode more complex characteristics such as goal proximity \cite{Nav:Navigating_by_Distance_Prediction}, semantic classes \cite{Nav:GoalOrientedSemanticExploration}, or goal reachability costs \cite{interactiveNav:Spatial-Action-Maps-for-Mobile-Manipulation}. Mostly these systems encode heuristics which align sub-optimally with the local planners objective, as they are optimized using different labels. The labels we use to train our global mapping module are generated on the fly from the same cost-map the local planner is optimized on. This way we achieve full consistency between global- and local path planning. For the mapping, we rely on traversability estimation, determining which parts of the visible terrain can be safely traversed. Geometry-based methods such as \cite{trav:geometric} using Lidar or depth cameras suffer from poor terrain estimates as deformable instances such as grass, shrubs, or snow cannot be reliably classified. Semantic-based methods such as \cite{trav:Schilling2017} successfully circumvent this problem by incorporating RGB information, but often suffer from a lack of robustness against appearance changes in real-world environments. In \cite{trav:WVN}, the authors utilize a visual transformer as a strong feature extractor and train a small network online for traversability adaption. We adopt this idea and fuse our representation learner with RGB information encoded by the pretrained DinoViT-V2 \cite{aux:DinoV2}. 

Various techniques in representation learning aim to extract the most relevant features from the input data. \cite{Representation:World_Models_Ha2018} explores this challenge by using a Variational Auto-Encoder (VAE) \cite{aux:VAE} in conjunction with an RNN to encourage the learning of a Latent Dynamics Model (LDM), showing that the RNN is able to learn a compact representation of the world that it can propagate forward in time. Similarly, \cite{Representation:Hoeller2021} applies this concept for dimensionality reduction, pre-training a world model to then train a policy network on the outputs of the world model using RL. Both works however rely on complicated training schedules with the different modules trained disjointly, leading to sub-optimal representations potentially containing information not relevant for the downstream task. We address this shortcoming by training our entire representation learner end-to-end. Using variational inference, we learn regularized representations of the environment, implicitly encoding information about the dynamic obstacles such as shape, position, or velocity.\\

\section{PLANNER ARCHITECTURE}

\begin{figure*}
    \centering
    \includegraphics[width=\linewidth]{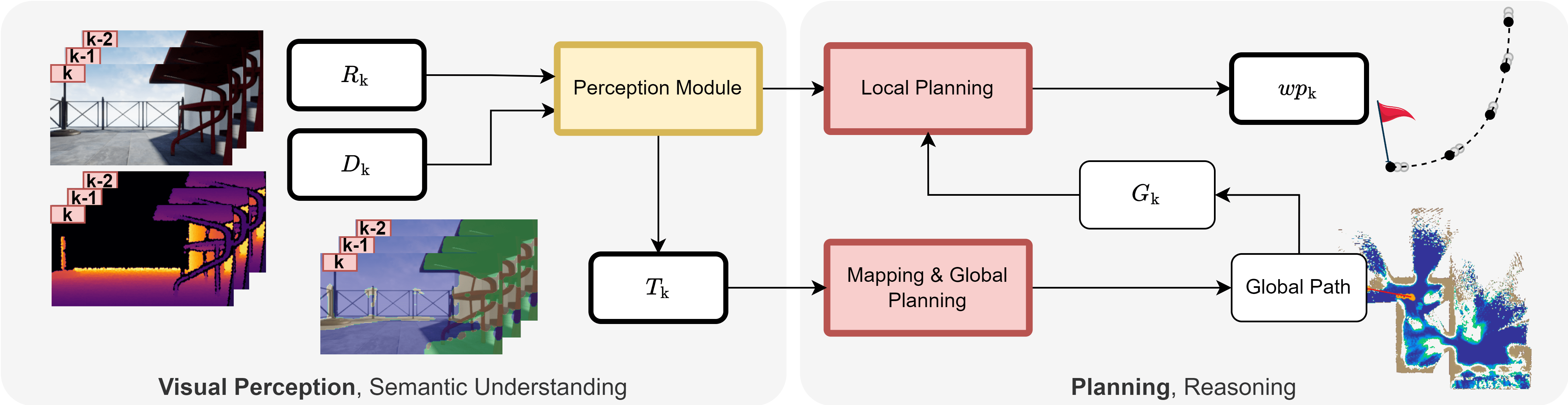}
    \caption{In each time step, RGB-D inputs ($R_\text{k}, D_\text{k}$) are provided to the planner. The perception module then produces traversability estimates $T_\text{k}$ which are continuously integrated into a map. Using this map, a global path to the goal is computed from which a subgoal $G_\text{k}$ is selected. The subgoal is fed to the local planner which computes the local path ($wp_\text{k}$). }
    \label{fig:network_architecture_overview}
\end{figure*}

\subsection{System Overview}
The global planner utilizes an online-generated global map to search for a potentially feasible path, which is then tracked by the local planner. The workflow of our planning algorithm is illustrated in Figure \ref{fig:network_architecture_overview}. Specifically, the perception module encodes the RGB observation $R_\text{k}$ and the depth observation $D_\text{k}$ at timestep $k$ into a compact representation $z_\text{k}$ and predicts a pixel-wise traversability mask $T_\text{k}$. The traversability mask is then fed to the mapping module where it is integrated into a 2D grid map. The grid map is used to compute a minimum-cost path to the goal by executing A* path search. From the discrete path, a subgoal is selected which is then fed to the local planner together with the embedding $z_\text{k}$. The local planner then predicts a path to the subgoal, consisting of $N$ intermediate waypoints. The idea is that the global planner uses explicit memory (grid-map) to capture the macroscopic structure of the environment and is therefore well suited to provide a long-term planning strategy. However, the semantic map has limited spatial resolution and contains temporary artifacts (decaying over time) in case of non-stationary obstacles. Therefore, we introduce a local planner that takes the responsibility of locally navigating around obstacles and deciding if an interaction is desirable. The local planner comprises a Recurrent Neural Network (RNN) which allows the planner to grasp short-term context from previous time steps. Combining both concepts enables the system to efficiently explore large environments, while locally considering interactions with obstacles to reduce the traversal effort. 

\subsection{Long-horizon planning}
The traversability mask $T$ encodes for each pixel whether it belongs to free space, a movable obstacle or non-traversable space. In order to integrate the mask into the map, we first compute the point cloud from the depth image using the pinhole projection model, then transform the points onto the world frame. We then project the points on the 2D plane and compute the respective grid cell index. Finally, we sample for each active grid cell the number of previous observations $m$ from the weight map to compute the update weight $\alpha$ (cf. Eq. \ref{eq:alpha}) and then update the grid map (cf. Eq. \ref{eq:cost_update}) in a weighted average fashion
\begin{eqnarray}
    \alpha &=& 2 / (1 + m) \label{eq:alpha}, \\
    c &=& \alpha \cdot c + (1 - \alpha) \cdot \Tilde{c}, \label{eq:cost_update}
\end{eqnarray}
where $c$ denotes the observed traversability class of the point and $\Tilde{c}$ denotes the average traversability class of the grid cell. The number of observations is clipped between $[0,100]$ such that the influence of new measurements does not eventually converge to zero. This leads to sufficiently fast decay of map-artifacts due to moving obstacles. For each updated grid cell the associated cell in the weight map is incremented. 

For global path planning, we resort to the A* algorithm with L2 heuristic. Due to the weighted averaging of the measurements during the mapping, our map no longer contains discrete area classes (0:free-space, 1:obstacle, 2:non-traversable, 3:unexplored) but a continuous belief value over the area classes. To obtain the discrete area classes back, we threshold the continuous values with the following ranges: $[0, 0.5)\rightarrow$ free space, $[0.5, 1.5)\rightarrow$ obstacles, $[1.5, 3.0)\rightarrow$ non-traversable, $[3]\rightarrow $ unexplored. We assign different penalties for planning through grid cells belonging to movable obstacles and non-traversable space. After the global path is found, we extract the element from the path in direct line of sight with the robot
we determine from the global path a set of points, in which each two consecutive points can be connected by a straight line without touching neither obstacle space nor non-traversable space. From this set, we use the closest point as the next subgoal $g_\text{k}$ for the local planner.

\subsection{Short-horizon planning}
\label{sec:local_planner}

\begin{figure*}[h]
    \centering
    \includegraphics[width=\linewidth]{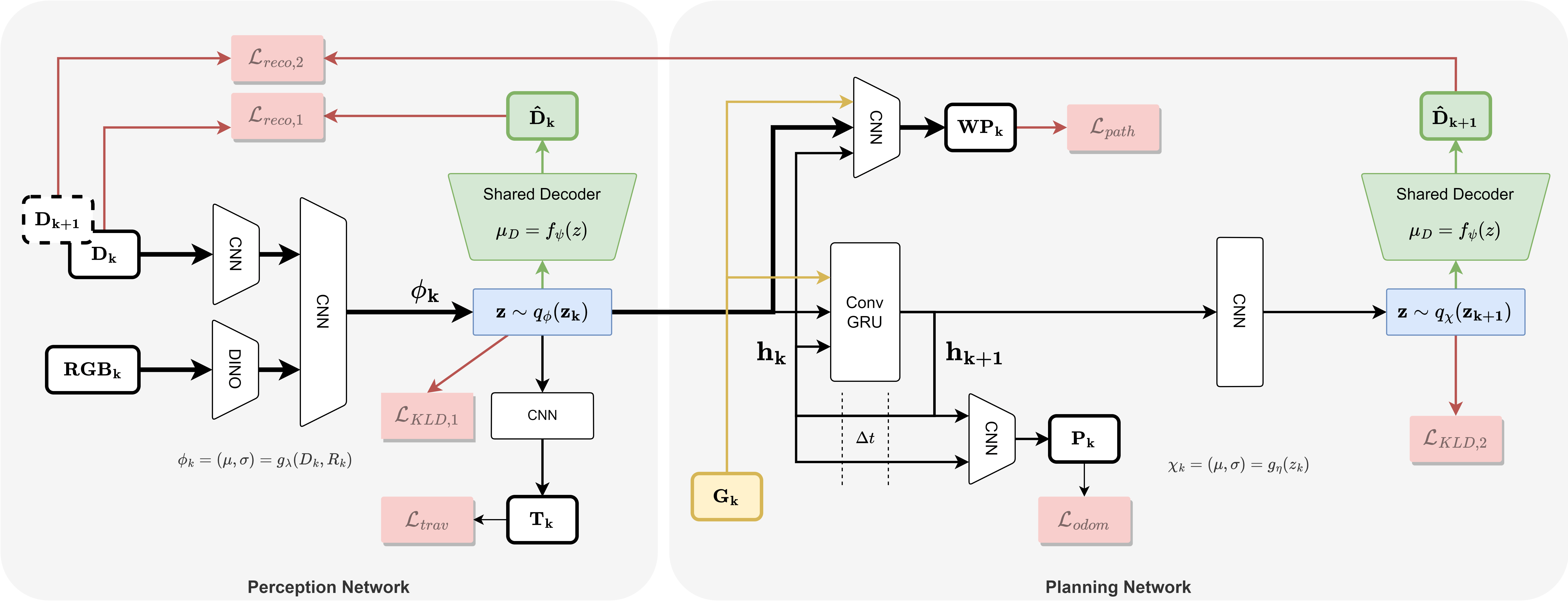}
    \caption{Learned modules of the planner, separated into perception (left) and planning (right). Trapezoids denote neural networks which change the spatial size of the embeddings. The blue boxes denote sampling blocks where an embedding tensor is drawn from a parametrized distribution $q_\phi(z)$. The residuals of the loss function are denoted by red boxes.}
    \label{fig:joint_implementation_overview}
\end{figure*}

Figure \ref{fig:joint_implementation_overview} illustrates our planner on an implementation level. The components can be categorized into perception (left) and planning (right).  We aim to learn a compact representation of the world from a sequence of visual observations, encoding relevant information about the environment such as position, velocity, and semantic information of surrounding obstacles. For this we let the perception network model the depth inputs $D$ by a small set of latent factors $z$, using a Latent Variable Model (LVM) of the form
\begin{align}
    \log p_{\boldsymbol{\theta}}(D) &= \log \int p_{\boldsymbol{\theta}}(D|z)p_{\boldsymbol{\theta}}(z)dz\\  
    &= \begin{aligned}[t]
        \mathbb{E}_{z\sim q_{\boldsymbol{\phi}}(z)} \bigg[\log \frac{p_{\boldsymbol{\theta}}(D|z)p_{\boldsymbol{\theta}}(z)}{q_{\boldsymbol{\phi}}(z)}\bigg] & \\
        + \underbrace{\mathbb{KL}(q_{\boldsymbol{\phi}}(z)||p_{\boldsymbol{\theta}}(z|x))}_{\geq 0}.
    \end{aligned}
    \label{eq:lvm_objective}
\end{align}
Reformulating the LVM using parameterized families of distributions we can resort to parameter optimization instead of variational calculus. Following \cite{aux:VAE} we choose the following parametrization
\begin{align}    
    p_{\boldsymbol{\theta}}(z) &\rightarrow \mathcal{N}(0,I) \label{eq:vae_prior}, \\    
    p_{\boldsymbol{\theta}}(D|z) &= \mathcal{N}(D| f_{\boldsymbol{\psi}}(z), I) = p_{\boldsymbol{\psi}}(D|z) \label{eq:vae_conditional}, \\    
    q_{\boldsymbol{\phi}}(z) &= \mathcal{N}(z|\boldsymbol{\phi}=\left(\mu, \Sigma\right)) \label{eq:vae_latent} \text{ with } \boldsymbol{\phi} = g_{\boldsymbol{\lambda}}([D,R]]),
\end{align}

where the parameters $\theta$ and $\phi$ are computed using armortized inference. For this the functions $g_\lambda$ and $f_\psi$ are learned from a dataset consisting of synthetic image sequences mixed with real-world images. To inject semantic information into the LVM, RGB information $R_\text{k}$ is used as an additional input for estimating the latent parameters $\phi$. We encode the RGB images using the DinoViT V2 \cite{aux:DinoV2} with frozen parameters, serving as a robust feature extractor. The depth input $D_\text{k}$ is encoded using a residual CNN. The inputs are then concatenated and fed through another residual CNN to estimate the latent parameters $\phi_\text{k}$. Both $\mu$ and $\Sigma$ are of shape $[23,40,8]$, thus we model the latent space with 7360 latent variables. We sample a latent tensor $z_\text{k} \sim q_{\phi_\text{k}}(z)$ and decode it using the shared-parameter transposed CNN $f_\psi$ to get a reconstruction $\hat{D}_\text{k}$ of the depth input $D_\text{k}$. Tracing the information flow from $D_\text{k}$ to $\hat{D}_\text{k}$ it becomes visible that the parameterized models involved in the computation form a Variational Auto-Encoder (VAE).
We also decode the mean tensor $\mu_\text{k}$ of the latent distribution to get the traversability mask $T_\text{k}$, using a transposed CNN. The traversability mask is used for the mapping as described earlier. The parameters of the latent distribution $\phi_\text{k} = [\mu, \Sigma]_\text{k}$ are passed to the convolutional GRU which additionally takes the subgoal $G_\text{k}$ as an input. The GRU propagates its hidden state $h_\text{k} \rightarrow h_\text{k+1}$. The mean of the latent $\mu_\text{k}$, the subgoal $G_\text{k}$, and the old hidden-state $h_\text{k}$ are fed to the Path-CNN which predicts $N$ waypoints $wp_\text{k}$, forming the local path to the subgoal. The new hidden-state $h_{\text{k}+1}$ is fed through the CNN $g_\eta(h_{\text{k}+1})$ to predict the parameters $\chi$ of the distribution over the next latent state $q_\chi(z_\text{k+1})$. We again sample a latent vector $z_\text{k+1} \sim q_\chi(z_\text{k+1})$ and pass it through the shared-parameter transposed CNN $f_\psi$. Tracing the information flow from $D_\text{k}$ to $\hat{D}_\text{k+1}$ it becomes visible that parameterized models involved in the computation form a predictive VAE.
Intuitively the GRU in the planner maintains a \textit{(world) belief state} $h$ and updates it based on the current \textit{(world) observation} $\phi_\text{k}$, resulting in the updated prediction of the \textit{belief state} $h_\text{k+1}$. We then use this prediction to infer the distribution over the next \textit{observation} $z_\text{k+1}$. With the structural bias of the RNN, our system is encouraged to learn the dynamics of the latent variables. As commonly done in LVMs, optimizing the Evidence Lower-Bound (ELBO, cf. Equation \ref{eq:lvm_objective}) introduces latent-regularization by minimization of the KL Divergence. Using an uncorrelated Unit-Normal prior encourages more disentangled representations eventually promoting decoupling in the learned system dynamics \cite{Representation:World_Models_Ha2018}\cite{Representation:Representation_Learning_for_Control_overview}. With this in mind, our side-objective can be understood as optimizing an LDM, which empirically has been shown to improve the expressiveness and robustness of the learned representations \cite{aux_task:AuxiliaryTasksSpeedUpLearning}.
As a second side-objective, we use the tensors $h_\text{k}$ and $h_\text{k+1}$ to predict the change in pose (position, heading) between the current and next time step, $P_\text{k}$. 

\section{TRAINING}
\label{sec:training}
\begin{figure}
    \centering
    \includegraphics[width=1\linewidth]{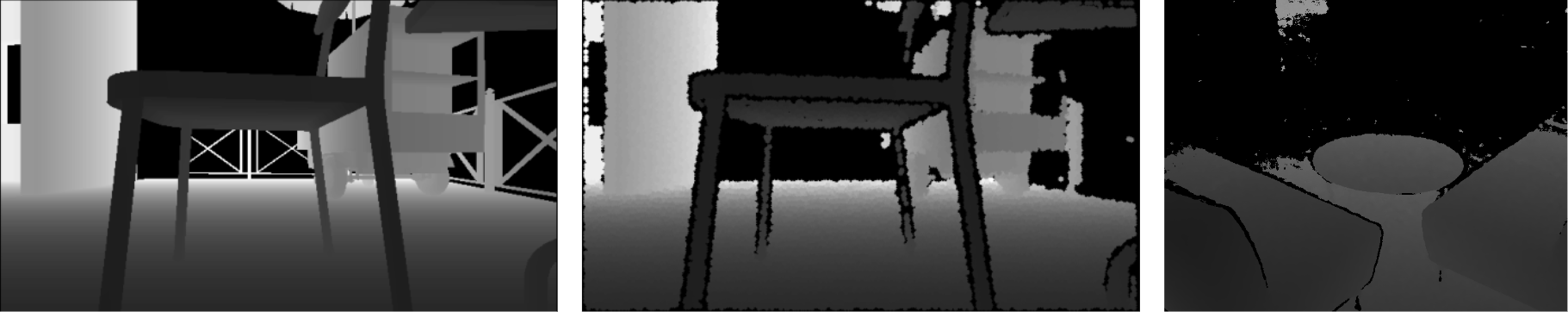}
    \caption{Ideal depth image from SPEAR (left) afflicted by simulated noise (mid) vs. Kinect depth image from the SUN-RGBD dataset (right).}
    \label{fig:noise_triplet}
\end{figure}

Our training set consists of sequential data, synthetically generated using Intel SPEAR. During data collection, the agent has access to the navigation mesh \cite{aux:recast_detour} encoding all traversal- and obstacle-interaction costs in the scene. This privileged information allows the agent to traverse along the minimum-cost path to the goal. Once the goal is reached, the environment resets. To demonstrate domain randomization, the positions and orientations of the agent and all obstacles are randomized on reset.
To account for sensor imperfections, we add white noise to the RGB images and simulate common depth sensor artifacts \cite{aux:depth_sensor_model} (cf. Figure \ref{fig:noise_triplet}). We accelerate the sampling procedure of \cite{aux:depth_sensor_model} using GPU vectorization (x500 speedup on i7-8750H w. RTX2070 mobile) allowing us to sample unique artifacts on a per-sample basis. This way, all noise artifacts are truly random between epochs. We also inject independent real-world images of multiple depth cameras from the SUN-RGBD dataset to robustify against unmodeled sensor intrinsics.

The training loss follows Figure \ref{fig:joint_implementation_overview}
\begin{eqnarray}
    \label{eq:lossweighting}
    \mathcal{L} &=& c_{nav} \cdot (c_{path} \cdot \mathcal{L}_{path} + \mathcal{L}_{odom}) + \nonumber \\
    && c_{trav} \cdot \mathcal{L}_{trav} + c_{repr} \cdot \mathcal{L}_{repr}
\end{eqnarray}
with
\begin{equation}
    \label{eq:lossweightingRepr}
    \mathcal{L}_{repr} = \mathcal{L}_{reco,1} + \mathcal{L}_{KLD,1} + \mathcal{L}_{reco,2} + \mathcal{L}_{KLD,2}.
\end{equation}
where the coefficients $c$ are hand-tuned parameters. The path loss follows
\begin{equation}
    \mathcal{L}_{PATH} = \alpha \cdot \mathcal{L}_{o} + (1-\alpha) \cdot (\mathcal{L}_{m} + \mathcal{L}_{g}),
\end{equation}

where $\mathcal{L}_{o}$ penalizes proximity towards obstacles, $\mathcal{L}_{m}$ penalizes irregular distances between consecutive waypoints and $\mathcal{L}_{g}$ penalizes a high L2 distance between last waypoint and goal. $\mathcal{L}_{o}$ is computed by projecting the waypoints onto the 2D costmap and sampling the cost values \cite{Nav:Iplanner}. The differentiable cost map (cf. Figure \ref{fig:method_overview}, Training) is derived from the navigation mesh, which we modify to encode obstacle locations and semantic classes (free space, light-weight / medium-weight / heavy-weight obstacle, non-traversable space) during simulation. The cost value of a grid-cell depends on the distance to the next obstacle and the obstacles semantic class. Using the tuning value $\alpha$ we manually balance the preference of the agent to be more goal-directed or more obstacle avoidant. 
$\mathcal{L}_{trav}$ is computed using the Multi-class Cross-Entropy. We compute the traversability labels by unprojecting the true depth image into a point-cloud, transforming it into world frame, and then projecting it onto the 2D grid map. We can then sample the cost values from the costmap and associate each cost value to each pixel from the depth image, such that we obtain the 2D ego-perspective traversability mask. 
For the auxiliary loss $\mathcal{L}_{odom}$ we take sum of the L2 norm of the translational error and the
L2 norm of the rotational error between the prediction of the change in pose and the actual change in pose.
We train the system on an NVIDIA RTX3090 for 10 epochs using PyTorch's AdamW optimizer with a constant learning rate of $1e-4$ and weight decay of $1e-8$. We also apply gradient scaling and train in half-precision (16 bit). We resort to a sequence length of $5$ time instances where each time instance has a batch of size $10$. Therefore, we use $50$ samples to compute the gradients. The DinoViT parameters are frozen. The dataset is shuffled after each epoch.

\section{EXPERIMENTS}


\begin{figure}
    \centering
    \includegraphics[width=\linewidth]{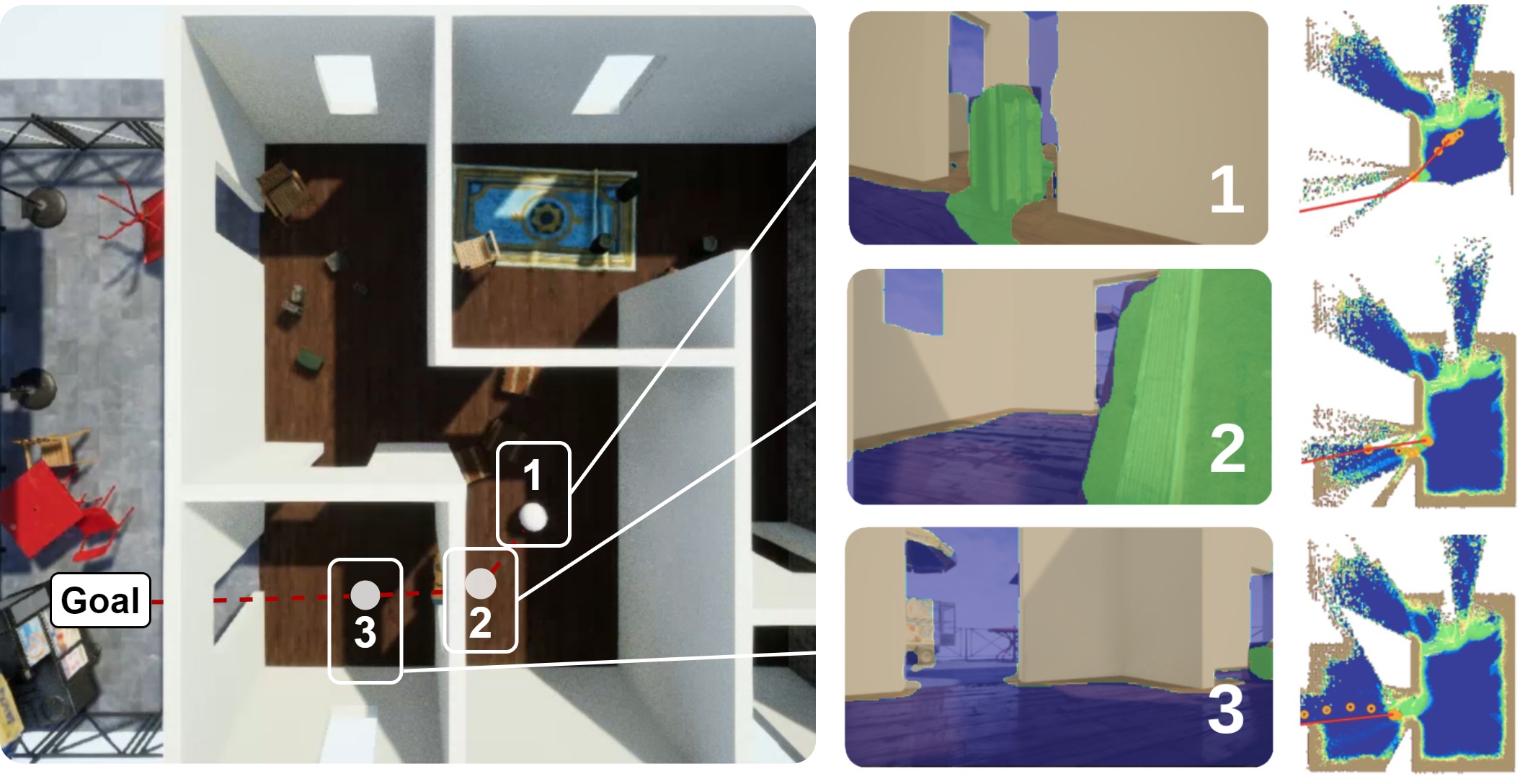}
    \caption{Left: A simulated agent (white sphere) approaches an obstacle (1, green), pushes it aside (2), and continues to the goal (3). Center: First-person traversability estimates overlaid on the RGB inputs. Right: The agent's global traversability grid map (blue: free space, green: interactive obstacles, brown: static obstacles).}
    \label{fig:interaction_demo_sim}
\end{figure}

We evaluate our planner across three distinct simulated environments: The maze-like \textit{House 1}, featuring manipulable obstacles that either appear in the dataset or are entirely new (cf. \ref{fig:interaction_demo_sim}); \textit{House 2}, populated exclusively with new obstacles that have textures different from those in the training set; and the cluttered \textit{Forest}, which contains no movable obstacles.

\subsection{Simulation Benchmarks}\label{simBenchmark}

We conduct each experiment over 100 epochs. At the start of each epoch, the agent is respawned, and we sample five random start/goal pairs, choosing the one with the greatest integral path length. This method eliminates overly simplistic start/goal pairs that could be solved with a direct path. An epoch is considered successful if the agent reaches its goal within 500 iterations (equivalent to 50 seconds at a rate of 10Hz). In our simulations, the agent navigates straight between local waypoints, executing in-place rotations upon reaching a target waypoint. An episode is deemed successful if the agent comes within 1 meter of the goal. Every 10 epochs, we reposition all non-stationary obstacles to a new location within a small area around their original, manually selected positions. For benchmarking, we employ the following metrics: the Success Rate (SR) (Equation \ref{eq:SR}) and Success Weighted by Path Length (SPL) (Equation \ref{eq:SPL})
\begin{eqnarray}
    \text{SR} &=& \frac{\text{num}_{\text{success}}}{\text{num}_{\text{success}} + \text{num}_{\text{failure}}} \label{eq:SR}\\
    \text{SPL} &=& \frac{1}{N} \sum_{i=0}^{N-1} S_i \frac{L_i}{\text{max}(L_i, \Tilde{L}_i)}. \label{eq:SPL}
\end{eqnarray}

where $L_i$ is the length of the traversed path and $\Tilde{L}_i$ is the length of the least-cost path computed with respect to the navigation mesh (only available in simulation, cf. \ref{sec:training}). We also capture the number of collisions with static elements $\eta_{\text{static}}$ and movable obstacles $\eta_{\text{obstacle}}$.
The performance of the \textit{proposed} planner, along with its ablated versions, is indicated in Table \ref{tab:ablation_learned_modules}. We observe that the agent manages to reach the goal in time with consistently high SR and SPL across all three test environments. We further observe that the agent successfully inherits the navigation preferences encoded into the costmap, as a result of optimizing the policy with respect to the $\mathcal{L}_{o}$ loss (cf \ref{sec:training}).
An example of interactive navigation is provided in Figure \ref{fig:interaction_demo_sim}, demonstrating the planner's ability to manage obstacles blocking the shortest path to the goal. In this scenario, the agent encounters a wooden crate on its direct route. It correctly identifies the crate as movable and determines that displacing the crate to proceed along the obstructed pathway is more resource-efficient than detouring through an alternative room. It then tips the crate over and proceeds with the traversal. This behavior demonstrates the planner's ability to intelligently negotiate obstacles, enhancing navigation strategies and, crucially, averting navigation failures by addressing obstructions directly when no unobstructed alternative paths are available.

\subsection{Ablation Study}
We ablate the planner to examine the impact of each design choice on the final planning performance. We also measure the performance of a simplistic \textit{baseline} configuration inspired by \cite{Nav:Bansal2019}, where we take the encoder architecture from the \textit{proposed} planner but instead predict the waypoints directly from the embeddings $\Phi_\text{k}$ using a convolutional residual network. All ablated configurations are trained under the same conditions as the \textit{proposed} planner, as outlined in Section \ref{sec:training}.

\label{sec:ablation}
The \textit{Deterministic} configuration has the stochastic blocks removed (blue blocks, Figure \ref{fig:joint_implementation_overview}) and is not optimized for the KL-Divergence. While having comparable SR and SPL to the proposed planner in the \textit{House 1} environment, it shows inferior performance on \textit{House 2} and \textit{Forest}. We also observe that the number of collisions $\eta_\text{static}$ increases for the scenes comprising a high obstacle diversity. This confirms the hypothesis that the latent regularization improves generalization by allowing for better interpolation in the latent space (cf. Section \ref{sec:local_planner}).

In the \textit{noRNN} configuration the local planner lacks memory. Thus, the network is not able to maintain a temporally consistent representation of the nearby obstacles. If an obstruction drops out of sight when circumventing, the agent turns prematurely towards the obstacle again. This reduces the path margin and increases the risk of a collision. Consequently, $\eta_\text{static}$ increases and the agent times out more often (reduced SR). 

In the \textit{onlyVAE} configuration, having both the RNN and predictive VAE branch removed, the SR significantly deteriorates in both the \textit{Forest} and \textit{House 2} environments, together with an increase in $\eta_\textit{static}$ in all test environments. Similar observations can be made when removing the odometry network (cf. \textit{noAUX}). This confirms the hypothesis that optimizing an LDM or odometry estimator as a co-objective benefits the learned representations and leads to improved generalization (cf. Section \ref{sec:local_planner}). 

The minimal \textit{Baseline} configuration performs equally good to the proposed planner in the less complex, non-interactive \textit{Forest} environment. However, in the interactive, labyrinth-like environments the performance drops significantly, underlining the efficacy of our architectural design choices.


\begin{table}[h]
    \centering
    \begin{tabular}{llcccccc}
        \toprule            
        Environment & Configuration & SR & SPL & $\eta_\text{obstacle}$ & $\eta_\text{static}$ \\
        \midrule
        \multirow{2}{*}{\textbf{House 1}} & Proposed        & \textbf{0.97} & 0.78 & 2.41 & \textbf{0.35} \\
                                          & Deterministic   & 0.97 & 0.76 & 2.14 & 0.55 \\
                                          & NoAUX           & 0.97 & 0.77 & 2.10 & 0.43 \\
                                          & NoRNN           & 0.92 & 0.72 & 1.99 & 0.66 \\
                                          & OnlyVAE         & 0.97 & 0.75 & 2.01 & 0.52 \\
                                          & Baseline        & 0.68 & 0.64 & 2.62 & 0.51 \\                                       
        \midrule
        \multirow{2}{*}{\textbf{House 2}} & Proposed        & \textbf{0.96} & 0.77 & 1.35 & \textbf{0.18} \\
                                         & Deterministic    & 0.62 & 0.77 & 2.10 & 0.86 \\ 
                                         & NoAUX            & 0.89 & 0.78 & 1.56 & 1.38 \\
                                         & NoRNN            & 0.86 & 0.75 & 1.37 & 0.42 \\
                                         & OnlyVAE          & 0.70 & 0.79 & 0.65 & 0.62 \\
                                         & Baseline         & 0.68 & 0.76 & 1.57 & 0.57 \\
        \midrule
        \multirow{2}{*}{\textbf{Forest}} & Proposed         & \textbf{0.96} & 0.74 & - & \textbf{0.14} \\
                                         & Deterministic    & 0.88 & 0.66 & - & 0.20 \\ 
                                         & NoAUX            & 0.84 & 0.70 & - & 0.74 \\
                                         & NoRNN            & 0.90 &	0.76 & - & 0.15 \\
                                         & OnlyVAE          & 0.86 & 0.65 & - & 0.40 \\
                                         & Baseline         & 0.92 & 0.71 & - & 0.29 \\
        \bottomrule
    \end{tabular}
    \caption{Quantitative evaluation of the planner and its ablated versions.}
    \label{tab:ablation_learned_modules}
\end{table}

\subsection{Real-World Experiments}

\begin{figure}
    \centering
    \includegraphics[width=1\linewidth]{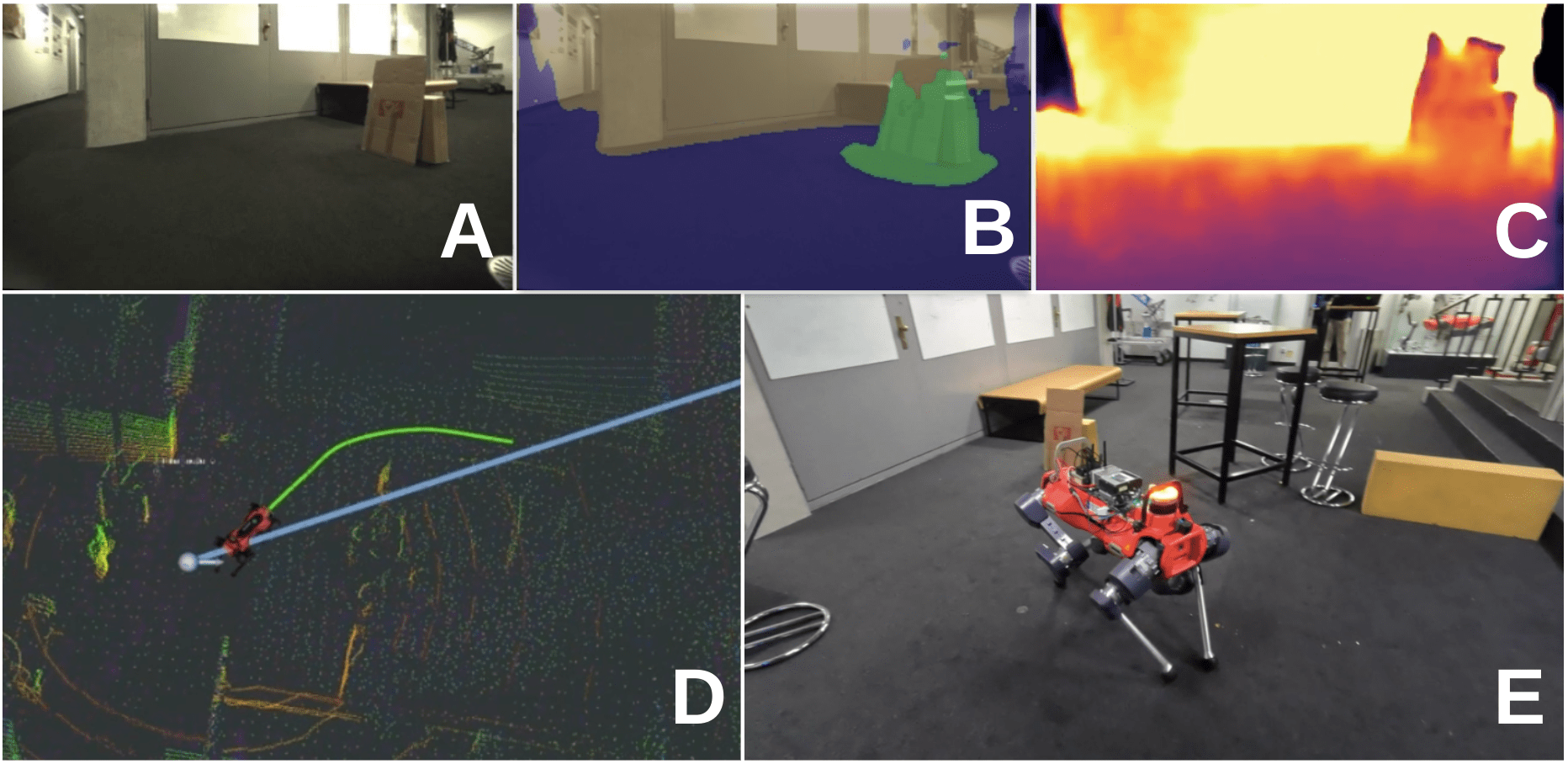}
    \caption{Planner anticipating to interact with obstacle (carton box) [E]. Also depicted: RBG input $R_\text{k}$ [A], traversability estimates $T_\text{k}$ overlaid on the RGB input [B], decoded depth $\hat{D}_\text{k}$ [C], local path prediction $wp_\text{k}$ visualized using RViz [D].}
    \label{fig:interaction_demo_real}
\end{figure}

\begin{figure}
    \centering
    \includegraphics[width=1\linewidth]{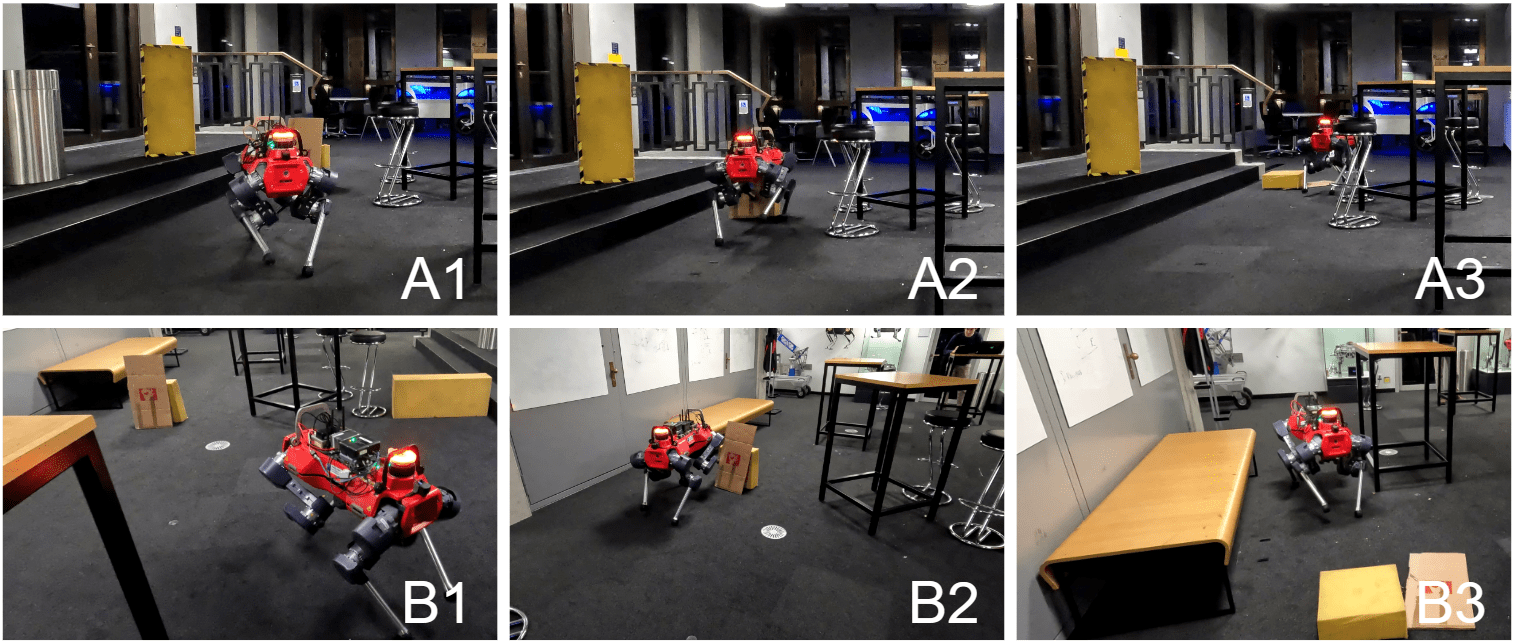}
    \caption{ANYmal negotiates obstructions which deny traversal along the shortest path to the goal.}
    \label{fig:realworld_demo}
\end{figure}

We test our planner at 5 Hz on the quadrupedal ANYbotics ANYmal D robot equipped with a NVidia Jetson AGX PC and an Intel Realsense D435 camera. The robot autonomously navigates from a start point to a predetermined goal without human intervention. We intentionally obstruct the direct route with unfamiliar obstacles not encountered during training (cf. Figure \ref{fig:interaction_demo_real} [A,E]), including both static (tables, heavy chairs, walls) and movable objects (paper boxes, lightweight office chairs, foam bricks). Our observations reveal that the planner effectively distinguishes between traversable and non-traversable areas (cf. Figure \ref{fig:interaction_demo_real} [B]) and consistently generates paths within the navigable space (cf. Figure \ref{fig:interaction_demo_real} [D]). It identifies movable obstacles (cf. Figure \ref{fig:realworld_demo} [A1,B1]), pushing them aside when bypassing is not feasible (cf. Figure \ref{fig:realworld_demo} [A2:3,B2:3]). Using the depth reconstructions (cf. Figure \ref{fig:interaction_demo_real} [C]) we can confirm that the obstacle shapes are successfully captured by the latent. From the local path visualization (cf. Figure \ref{fig:interaction_demo_real} [D]) we can observe that the local path correctly bends around static obstacles, confirming that the learned planner correctly reasons about obstacles. Moreover, the stability of local path predictions over time, even when obstacles are no longer visible, underscores the planner's temporal consistency.

\section{CONCLUSION and FUTURE WORK}
We presented IN-Sight: a visual path planner for interactive navigation in cluttered and maze-like environments, demonstrating successful transfer to robotic hardware. We introduced a self-supervised training approach, leveraging the navigation mesh to auto-generate consistent path- and traversability supervision. This significantly reduces the manual labor required to create large photorealistic datasets suitable for planner and traversability estimator training, opening new opportunities for interactive navigation research. For future work, we aim to enhance our current setup in terms of sim-to-real transfer. By using low Level of Detail (LOD) collision meshes for the obstacles, we can generate depth reconstruction labels $\hat{D}_\text{k}$ with low LOD. These labels are expected to facilitate the learning of more basic structural representations, thereby improving the system's ability to generalize. Different from that, we seek to explore the application of Large Vision-Language Models for reasoning about the movability of obstacles. Utilization of the extensive amount of data encapsulated within these foundation models combined with their ability to reason, could potentially allow for the consideration of minute details in the environment, leading to more accurate predictions of whether an obstacle can be displaced.


\addtolength{\textheight}{-12cm}  




\section*{ACKNOWLEDGMENT}
This work was supported by Intel Labs Munich, in a collaboration between the Robotic Systems Lab at ETH Zurich, and the Smart Robotics Lab at TUM. Moreover, this work has been conducted as part of ANYmal Research, a community to advance legged robotics.


\bibliographystyle{bibliography/IEEEtranN}
\bibliography{bibliography/references}
\addcontentsline{toc}{chapter}{Bibliography}

\end{document}